\documentclass[conference]{IEEEtran}
\IEEEoverridecommandlockouts
\usepackage{cite}
\usepackage{amsmath,amssymb,amsfonts}
\usepackage{algorithmic}
\usepackage{graphicx}
\usepackage{textcomp}
\usepackage{xcolor}
\usepackage{verbatim} 
\usepackage{multirow}
\usepackage[inline]{enumitem}
\usepackage{tabularx}
\usepackage{balance}
\usepackage{subcaption}
\usepackage{booktabs}
\usepackage{algorithm}
\usepackage{hyperref}
\usepackage{xfp}

\usepackage[labelformat=parens,labelsep=quad,skip=3pt]{caption}

\def\BibTeX{{\rm B\kern-.05em{\sc i\kern-.025em b}\kern-.08em
		T\kern-.1667em\lower.7ex\hbox{E}\kern-.125emX}}
		
	\makeatletter

\def\ps@IEEEtitlepagestyle{%
  \def\@oddfoot{\mycopyrightnotice}%
  \def\@evenfoot{}%
}
\def\mycopyrightnotice{%
  {\footnotesize 978-1-7281-9556-8/21/\$31.00 \textcopyright2021 IEEE \hfill}
  \gdef\mycopyrightnotice{}
}

\begin{document}

\title{Compact CNN Models for On-device Ocular-based User Recognition in Mobile Devices}

\author{Ali Almadan and Ajita Rattani\\
School of Computing\\
Wichita State University, USA\\
{\tt\small aaalmadan@shockers.wichita.edu, ajita.rattani@wichita.edu}
}

\maketitle
\thispagestyle{empty}

\begin{abstract}
A number of studies have demonstrated the efficacy of deep learning convolutional neural network (CNN) models for ocular-based user recognition in mobile devices. However, these high-performing networks have enormous space and computational complexity due to the millions of parameters and computations involved. These requirements make the deployment of deep learning models to resource-constrained mobile devices challenging. To this end, only a handful of studies based on knowledge distillation and patch-based models have been proposed to obtain compact size CNN models for ocular recognition in the mobile environment. In order to further advance the state-of-the-art, this study for the first time evaluates five neural network pruning methods and compares them with the knowledge distillation method for on-device CNN inference and mobile user verification using ocular images. Subject-independent analysis on VISOB and UPFR-Periocular datasets suggest the efficacy of layerwise magnitude-based pruning at a compression rate of \textbf{8} for mobile ocular-based authentication using ResNet-50 as the base model. Further, comparison with the knowledge distillation suggests the efficacy of knowledge distillation over pruning methods in terms of verification accuracy and the real-time inference measured as deep feature extraction time on five mobile devices, namely, iPhone 6, iPhone X, iPhone XR, iPad Air 2 and iPad 7th Generation.
 \end{abstract}
\section{Introduction}
Ocular biometrics consists of regions in the eye and those around it, i.e., iris, conjunctival vasculature, and periocular region. Ocular biometrics has obtained significant attention from the research community and industry alike due to its accuracy, security, better robustness against facial expressions, and ease of use in mobile devices~\cite{kumari2019periocular,Reddy2018,de2015mobile}. An ocular region can also be scanned in the presence of facial masks worn extensively in pandemics such as in COVID-19. 
Further, visible light ocular biometric modality can be acquired using almost any imaging device ranging from high-end DSLR to ubiquitous mobile device RGB cameras. Ocular biometrics in the RGB spectrum have been utilized for smartphone-based login authentication by both academia and industry such as EyeVerify Inc. In fact, mobile ocular-based user authentication can operate in darker environments by using the device screen as a light source~\cite{reddy2020generalizable}. 

A number of methods based on hand-crafted textural descriptors such as histograms of oriented gradients (HOG), local binary patterns (LBP), local phase quantization (LPQ), and binary statistical image features (BSIF) have been used for person authentication using ocular images~\cite{rattani2016icip} in mobile devices. With advances in deep learning, deeply coupled autoencoders and different convolutional neural network (CNN) architectures such as ResNet-50, and MobileNet have been fine-tuned for ocular-based user recognition (authentication) in mobile devices~\cite{proencca2017deep,VISOB,RAJA2020103979,rattani2017fine}. 

The challenges of unconstrained mobile environments may lead to substantial variations in the ocular samples due to factors such as lighting conditions, distance, motion blur, glasses and occlusion due to hair, front-facing imaging sensor and optic aberrations (including smudged lenses), and other imaging issues such as inaccurate white balance and exposure metering. These factors result in performance degradation in the mobile environment. Datasets such as VISOB~\cite{Rattani2016,VISOB} and UFPR-Periocular~\cite{zanlorensi2020ufprperiocular} have been assembled for research and development in ocular-based user authentication\footnote{The term authentication and recognition is used interchangeably in this paper.} in the mobile environment. 

Most of the earlier ocular recognition studies in the mobile environment used CNNs under closed-set subject-dependent evaluation protocols, where the subjects overlap between the training and test sets, which may overestimate performance and generalizability~\cite{rattani2017fine,rattani2016icip}. But recently, researchers have been focusing on CNN-based methods for ocular recognition under the subject-independent protocol~\cite{reddy2020generalizable,VISOB}. The subject-independent protocol means the subjects do not overlap between the training and test set. Therefore, the subject-independent analysis, although more challenging, is much better for developing and testing generalizable models compared to the closed-set subject-dependent analysis. Further, the need to re-train the CNN model every time a new subject is enrolled (the scalability issue) is mitigated by the subject-independent protocol.

These high-performing deep CNN models have enormous space and computational complexity due to the millions of parameters and computations involved~\cite{8763885}. The number of parameters is dependent on the learnable layers i.e., convolution and fully connected layers. The \emph{size} of the model increases with the number of parameters in the learnable layers. The operations such as convolution and matrix multiplication are built upon multiply-add (MAdd) operations which constitute the parameters of the model. 
In a convolution layer, let $W\times H$ be input spatial resolution for a layer with $ch_{in}$ feature channels and $K\times K$ convolutional kernels to generate $ch_{out}$ feature channels, then the number of MAdd operations is given as:
\begin{equation}
W\: \times\: H\: \times\: ch_{in}\: \times\: ch_{out} \times\: K\: \times\: K
\label{eq:one}
\end{equation}
Further, In a fully connected layer with $F_{in}$ number of input features and $F_{out}$ number of output features, the number of MAdd operations are given as 
%
%
\begin{equation}
F_{in} \times\: F_{out}
\label{eq:two}
\end{equation} 

These MAdd operations represent the \emph{parameters} and \emph{computational cost} of the model. 

The high computational cost of large deep CNNs makes the deployment to resource-constrained mobile environment challenging~\cite{8763885}. 
The size and computational cost of the deep learning models should be low for real-time and frequent on-device biometric authentication on mobile devices, for instance, in case of unlocking the phone. This would provide an enhanced \emph{user experience} due to faster authentication mechanism as high-cost models are slower to execute. Also, this would give \emph{battery-friendly biometric authentication}. 

Methods based on network pruning~\cite{blalock2020state}, lower bit quantization~\cite{choukroun2019lowbit}, knowledge distillation~\cite{Gou_2021}, squeezed convolutional networks, and the use of separable convolutions~\cite{sandler2018mobilenetv2,ma2018shufflenet} have been introduced to reduce the size and computational cost of the models.

A handful of studies have proposed lightweight (compact size) CNN models for ocular recognition in the mobile environment. These studies implemented knowledge distillation~\cite{Gou_2021}, an ensemble of patch-based ocular CNNs~\cite{Reddy2018} and a customized version of MobileNet-V2~\cite{reddy2020generalizable} for obtaining lightweight models for ocular based user authentication in smartphones. However, the state-of-the-art is still in its initial stages for compact size models for ocular-based mobile user authentication. In order to further advance the state-of-the-art, the contributions of this work are as follows:

\begin{itemize}
    \item Evaluation of the state-of-the-art neural network pruning models~\cite{blalock2020state} namely; Global Magnitude Pruning, Layerwise Magnitude Pruning, Global Gradient Magnitude Pruning, Layerwise Gradient Magnitude Pruning along with Random Pruning using ResNet-50 as base architecture fine-tuned for ocular-based user authentication in smartphones.
    \item Comparison with the compressed version of ResNet-50 models, namely; ResNet-20 and ResNet-8 along with lightweight MobileNet-V2 and ShuffleNet-V2 models, trained using knowledge distillation~\cite{Gou_2021}.
    \item Subject independent evaluation of all the compact lightweight models on  VISOB~\cite{VISOB}  and UFPR-Periocular~\cite{zanlorensi2020ufprperiocular} on mobile ocular datasets.
    \item Inference time evaluation (in terms of deep feature extraction time) of all the compressed models by real-time implementation on iPhone 6, iPhone X, iPhone XR, iPad Air 2, and iPad 7th generation.
\end{itemize}

This paper is organized as follows: Section 2 discusses the prior work on compact models for ocular recognition in mobile devices. Section 3 discusses the pruning and knowledge distillation techniques used in this study. Section 4 elaborates on the CNN architecture considered. Datasets and experimental protocol are discussed in section 5. Results are discussed in section 6. Conclusions are drawn in section 7.

\section{Prior Work}
In this section, we discuss the existing studies on developing lightweight CNN models for ocular-based user recognition in smart-phones.

Boutros et al.~\cite{9210990} proposed a lightweight DenseNet-20 deep
learning model with only $1.1$m trainable parameters~\cite{9210990} obtained via knowledge distillation. Experiments performed on the VISPI dataset showed that DenseNet-20 trained using knowledge distillation outperforms the same model trained without knowledge distillation with EER reduction from $8.36\%$ to $4.56\%$.

Reddy et al.~\cite{Reddy2018} proposed patch-based OcularNet, a convolutional neural network (CNN) model, that used patches from the eye images for user recognition. For the OcularNet model, six registered overlapping patches were extracted from the ocular region, and a small convolutional neural network (CNN) was trained for each patch to extract feature descriptors. The verification performance of the proposed OcularNet which has $1.5$M parameters was compared to the popular ResNet-50 model which has $23.4$M parameters. When trained on the VISOB dataset, the proposed OcularNet model obtained equivalent performance over ResNet-50 in the subject independent verification setting. In another study~\cite{reddy2020generalizable}, authors proposed a customized version of the MobileNet-v2 architecture~\cite{reddy2020generalizable}
obtained by removing the last convolutional layers from the original implementation without affecting the accuracy while reducing the model size by $3.4\times$ compared to the original MobileNet-v2, and $36\times$ compared to the popular ResNet-50 for mobile ocular recognition in subject-independent analysis.


Jung et al.~\cite{jung2020periocular} proposed the transfer of information from the face to periocular modality by means of knowledge distillation (KD) for feature learning. However, the authors reported that applying typical KD techniques to heterogeneous modalities directly is sub-optimal. 


\section{Model Compression Techniques}
Several approaches have been proposed for compression of deep learning networks. In this study, we evaluate pruning and knowledge distillation techniques for obtaining lightweight (compact) models. 

\begin{figure}
    \centering
    \includegraphics[width=0.48\textwidth]{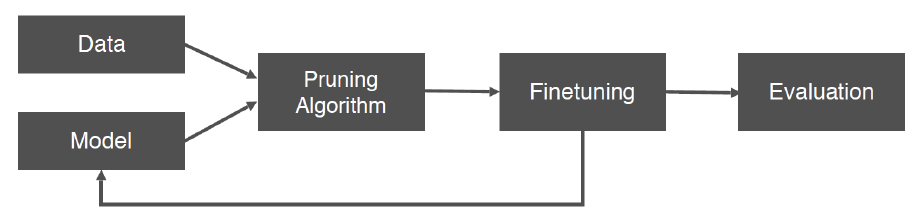}
    \caption{The steps involved in pruning a neural network architecture.}
    \label{prune}
\end{figure}

\subsection{Network Pruning}
\label{sec:prune}
Neural network pruning is the task of \emph{reducing the size of a network} by removing parameters. This entails systematically removing parameters from an existing network~\cite{blalock2020state}. Typically, the initial network is large and accurate, and the goal is to produce a smaller network with similar accuracy. There are several methods of obtaining a pruned model mainly derived from Algorithm 1. The network is first trained to convergence. Afterward, each parameter or structural element in the network is issued a score, and the network is pruned based on these scores. Pruning reduces the accuracy of the network, so it is trained further (known as fine-tuning) to recover. The process of pruning and fine-tuning is often iterated several times, gradually reducing the network’s size as shown in Figure~\ref{prune} and Algorithm~\ref{alg:prune_algorithm}. 
In Algorithm~\ref{alg:prune_algorithm}, $f(X;W)$ represent the neural network model to be pruned.

\begin{algorithm}[h]
\caption{Iterative Pruning with Fine-Tuning}
\label{alg:prune_algorithm}
\begin{algorithmic}[1]
\REQUIRE $N$, the number of iterations of pruning, and \\ \hspace{1.5em}$X$, the dataset on which to train and fine-tune

    \STATE $W \gets initialize()$
    \STATE $W \gets trainToConvergence(f(X; W))$
    \STATE $M \gets 1^{|W|}$
    \FOR{$i$ \text{ }in $1$ to $N$}%
        \STATE $M \gets prune(M, score(W))$%
        \STATE $W \gets fineTune(f(X; M \odot W))$%
    \ENDFOR
    \STATE \textbf{return} $M, W$
\end{algorithmic}
\end{algorithm}

These pruning methods differ based on whether they prune individual parameters (i.e., unstructured pruning). Other methods consider parameters in groups (structured pruning), removing entire layers or channels to exploit hardware and software optimized for dense computation. Some pruning methods compare scores locally, pruning a fraction of the parameters with the lowest scores within each structural sub-component of the network (e.g., layers). Others consider scores globally, comparing scores to one another irrespective of the part of the network in which the parameter resides~\cite{blalock2020state}.



\par In this study, we evaluated five existing pruning methods~\cite{blalock2020state} described as follows:
\begin{itemize}

\item \textbf{Global Magnitude Pruning} - prunes the weights with the lowest absolute value anywhere in the network.

\item \textbf{Layerwise Magnitude Pruning}  - for each layer, prunes the weights with the lowest absolute value.

\item \textbf{Global Gradient Magnitude Pruning} - prunes the weights with the lowest absolute value of ($weight \times gradient$), evaluated on a batch of inputs.

\item \textbf{Layerwise Gradient Magnitude Pruning}- for each layer, prunes the weights with the lowest absolute value of ($weight \times gradient$), evaluated on a batch of inputs.

\item \textbf{Random Pruning} - prunes each weight independently with the probability equal to the fraction of the network to be pruned.
\end{itemize}
The pruning metric is known as the Compression Ratio (CR) which is defined as:
$\frac{\text{original size}}{\text{compressed size}}$
\begin{figure}
    \centering
    \includegraphics[width=0.48\textwidth]{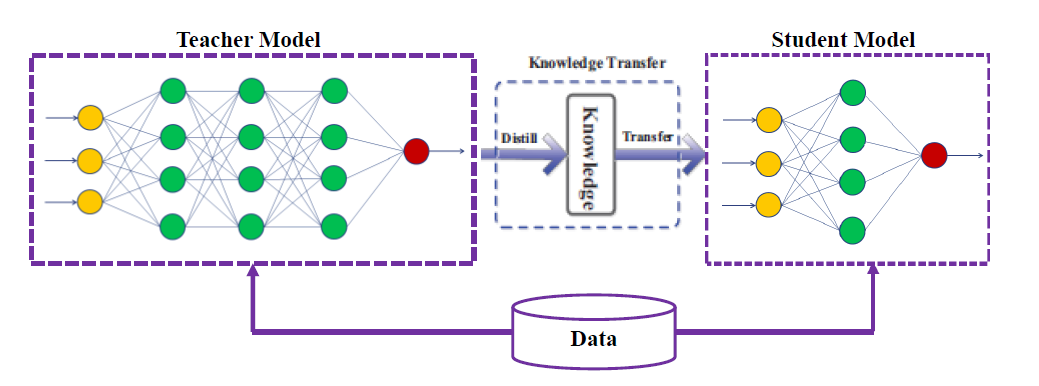}
    \caption{Generic architecture of knowledge distillation using a teacher-student model.}
    \label{fig:KD}
\end{figure}

\subsection{Knowledge Distillation}
Knowledge Distillation (KD)~\cite{Gou_2021} is a technique to improve the performance and generalization ability of smaller models by transferring the knowledge learned by a cumbersome model (teacher) to a single small model (student). \emph{The key idea is to guide the student model to learn the relationship between different classes discovered by the teacher model that contains more information beyond the ground truth labels}. 

To be specific, given a labelled data set \{xi, yi\}, i= 1,…n 
where $x_i$ is the data and $y_i$ is the label,
the knowledge from the teacher’s prediction
$p_\tau^T$ is distilled to the student’s prediction $p_\tau^S$ by minimizing the loss function as follow:
\begin{equation}
     L = (1-\lambda) L_{ce} + \tau^2 \lambda L_{kD},
\end{equation}

where $L_{ce}=-yi\sum_i log p(yi|xi)$ is the cross-entropy loss with student network prediction $p(yi|xi)=softmax f(xi)$
and f(.) is the logit of the network. $L_{KD}$ is the KD loss defined as:

\begin{equation}
     L_{KD} = \sum_i KL(p_\tau^T(y_i|x_i)||p_\tau^S(y|x_i))
\end{equation}

Here, $p_\tau(y|x)$ = softmax $(f(x)/\tau)$ and $p_\tau(y|x)$ = softmax$(f(x)/\tau)$ are the smoothen student’s prediction and teacher’s prediction, respectively.
The smoothness of the prediction is regulated by the
temperature term $\tau$ and KL is defined as the KL divergence. The hyperparameter $\lambda$ decides the contribution of the KD loss. Figure~\ref{fig:KD} shows the generic architecture of the knowledge distillation using teacher-student model. We used knowledge distillation method described above to train the compact version of ResNet-50 model and other lightweight mobile friendly models for mobile ocular recognition.

\section{CNN Architectures}
In this section, we will describe the CNN architectures used as a base model for compression or as a student model for knowledge distillation.
\begin{figure}
    \centering
    \includegraphics[width=0.20\textwidth]{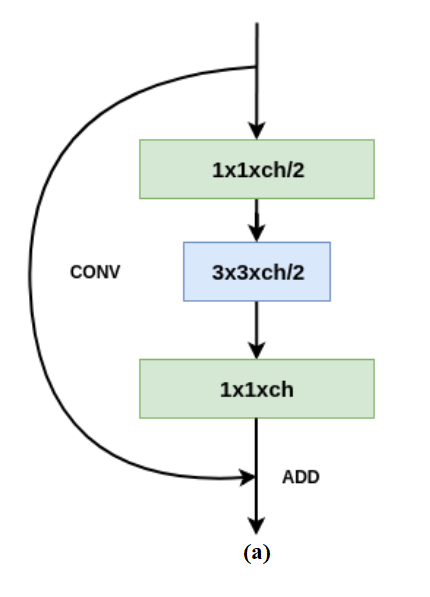}
    \includegraphics[width=0.22\textwidth]{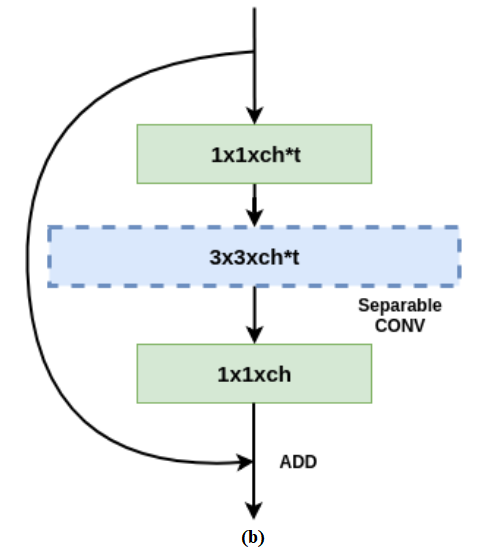}
    \includegraphics[width=0.13\textwidth]{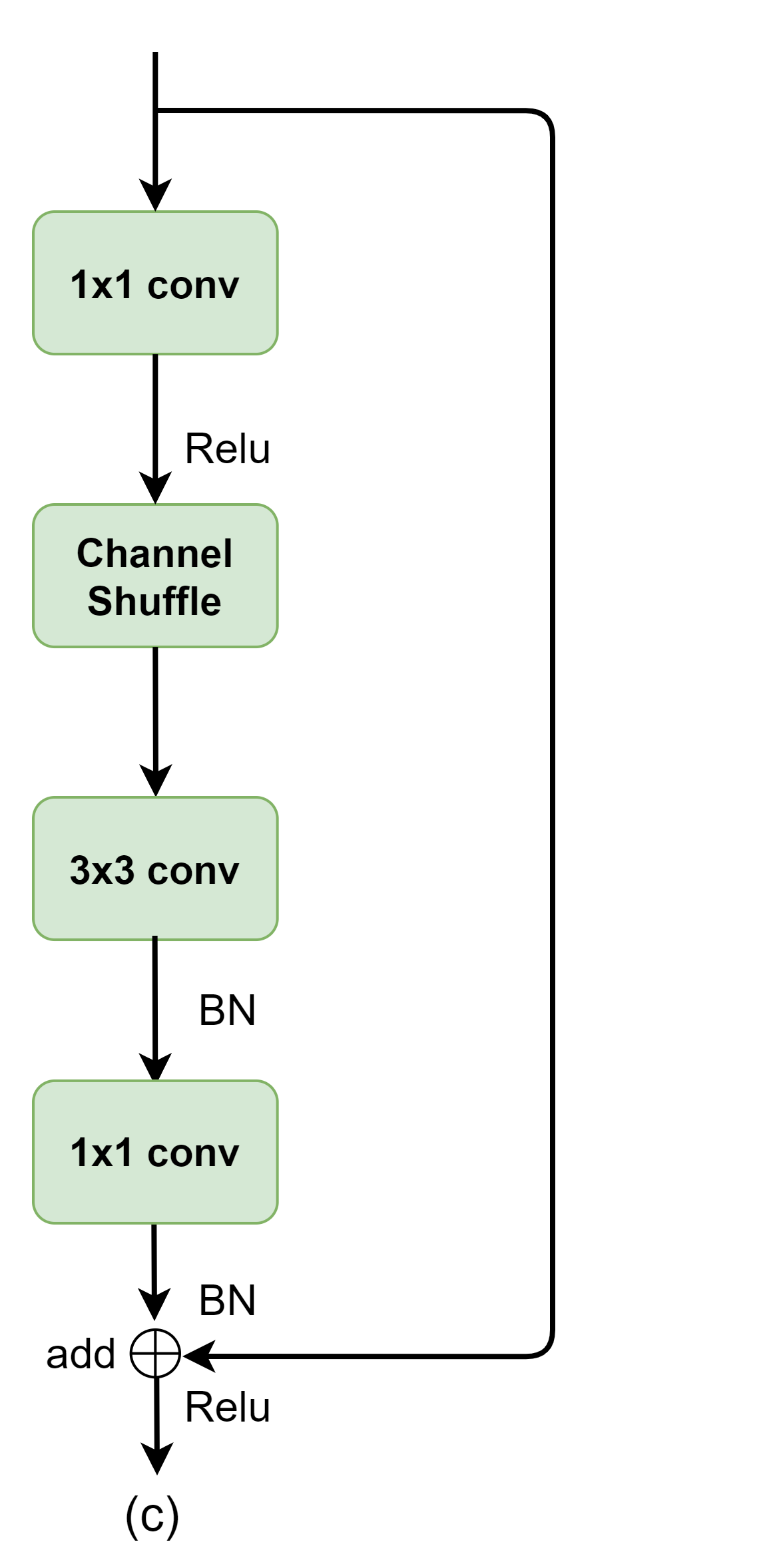}

    \caption{Architecture of (a) ResNet, (b) MobileNet, and (c) ShuffleNet CNN models.}
    \label{fig:cnn}
\end{figure}

\begin{itemize}
    \item \textbf{ResNet-50}: ResNet~\cite{he2016deep} is a short form of residual network based on the idea of “identity shortcut connection” where input features may skip certain layers (Figure~\ref{fig:cnn}(a)). The residual or shortcut connections introduced in ResNet allow for identity mappings to propagate around multiple nonlinear layers, preconditioning the optimization and alleviating the vanishing gradient problem. The ResNet-50 is used as the heavyweight model pruned using the methods discussed in section 3.1 and as a teacher for knowledge distillation mentioned in section 3.2. Other variations of ResNet consisting of eight and twenty residual layers, ResNet-8 and ResNet-20, respectively, were used as student models in this study.
    
    \item \textbf{MobileNet}: MobileNet-V2~\cite{sandler2018mobilenetv2} is one of the most popular mobile-centric deep learning architectures which is small in size as well as computationally efficient. The main idea of MobileNet is that instead of using regular $3\times3$ convolution filters, the operation is split into depth-wise separable $3\times3$ convolution filters followed by $1\times 1$ convolutions. While achieving the same filtering and combination process as a regular convolution, the new architecture requires a fewer number of operations and parameters (Figure~\ref{fig:cnn}(b)). MobileNet-V3-Small~\cite{howard2019searching} is a variation of MobileNets. It is utilized to run on mobile phone CPUs targeting low-resource constraint devices. MobileNet-V2 and V3 are both used as student models in this study.
    
    \item \textbf{ShuffleNet}: ShuffleNet-V2-50~\cite{ma2018shufflenet} is a variation of ShuffleNet which is a mobile-centric deep learning model.  It builds upon ShuffleNet-v1, which utilized pointwise group convolutions, bottleneck-like structures, and a channel shuffle operation. However, the direct metric, e.g., speed, also depends on the other factors such as memory access cost and platform characteristics (Figure~\ref{fig:cnn}(c)). ShuffleNet-V2-50 is used as a student model in this study.
    
\end{itemize}

\section{Datasets and Protocol}
In this section, we discuss datasets used for training and evaluating the compressed models. 
\subsection{Datasets}
\begin{itemize}
    \item \textbf{VISOB 2.0~\cite{VISOB}:} VISOB $2.0$ is the 2nd version of VISOB 1.0 dataset used in IEEE WCCI competition 2020 that facilitates subject-independent analysis. This publicly available dataset consists of a stack of eye images captured using the burst mode via two mobile devices:  Samsung Note 4  and OPPO  N1. During the data collection,  the volunteers were asked to take their selfie images in two visits, 2 to 4 weeks apart from each other. At each visit, the selfie-like images were captured using the front-facing camera of the mobile devices under three lighting conditions  (daylight,  office light,  and dark light) and two sessions (about $10$  to $15$  minutes apart). The stack consisting of five consecutive eye images was extracted from the stack of full-face frames selected such that the correlation coefficient between the center frame and the remaining four images is greater than $90\%$.
    
    \par This dataset was used for fine-tuning the ResNet-50 models and pruned networks. We used $40K$ images captured from $200$ subjects from VISOB 2.0 captured from all devices and different lighting conditions for CNN models training/ fine-tuning. Images of left and flipped versions of right eyes were considered to be an individual subject. 
    \par The remaining $350$ subjects were used for subject-independent verification performance evaluation of the compressed models for mobile ocular recognition. 
    
    \item \textbf{UFPR-Periocular~\cite{zanlorensi2020ufprperiocular}:} 
    This is the latest periocular biometric dataset containing samples from $1,122$ subjects with a total of $33,660$ images acquired in 3 sessions by $196$ different mobile devices. The data is captured across race, age, and gender. The gender distribution of the subjects is ($53.65\%$) male and ($46.35\%$) female, and approximately $66\%$ of the subjects are under $31$ years old. The eye corners of the ocular images are annotated with $4$ points per image (inside and outside eye corners), to normalize the periocular region across scale and rotation. This dataset is used for the subject-independent evaluation of the compressed models.
    
        \end{itemize}

\subsection{Protocol}
	
The ResNet-50 model fine-tuned on $200$ subjects from VISOB 2.0 dataset~\cite{VISOB} was used as a baseline model to be pruned and a teacher for knowledge distillation experiments. For the purpose of experiments, all the images were resized to $224\times 224$. After the last convolutional layer, we added batch Normalization (BN), dropout, the fully connected layer of size 512, and a batch normalization followed by the final output layer. This would result in a 512-D feature embedding. The ResNet-50 model was trained using an early stopping mechanism using Adam optimizer and a cross-entropy loss function.

\par The open-set evaluation is performed on the pruned version of the ResNet-50 model and the trained student networks (ResNet-8, ResNet-20, MobileNet-V2, MobileNet-V3, and ShuffleNet-V2). The template set consists of a stack of $5$ ocular images per subject for VISOB and UFPR across two different sessions. For VISOB, the template and test sets consist of multiple stacks of $5$ images per individual. The UFPR dataset has two stacks of $5$ images: one used for the template and another as the test set.  The matching score between deep features extracted from a pair of template and test image is calculated using the cosine similarity metric given in eq~\ref{cos_sim}:

\begin{equation}
\label{cos_sim}
    	\operatorname{sim}(u, v)
= \frac{u \cdot v}{|u||v|}
= \frac{\sum_{i = 1}^N u_i v_i}
{\sqrt{\left(\sum_{i = 1}^N u_i^2\right)
		\left(\sum_{i = 1}^N v_i^2\right)}}
\end{equation}
	where $u$ and $v$ are the two deep feature vectors: $u = \{u_1,u_2,\dots,u_N\}$ 
	and $v = \{v_1,v_2,\dots,v_N\}$.
\par Cosine similarity metrics is widely adopted for measuring the angles between deep features extracted from a pair of images~\cite{liu2017sphereface}.
\par We trained our student models using Stochastic Gradient Descent (SGD) optimizer with a batch size of $32$. MobileNet-V2 was trained for 50 epochs, ShuffleNet for 100 epochs, ResNet-20 and ResNet-8 for 200 epochs, and MobileNet-V3 for 35 epochs using an early stopping mechanism based on the validation set accuracy (indicated in Table~\ref{tab:kd_results}). All trained models were fine-tuned on ImageNet and Softmax used as a loss function during the training process.
All pruning methods were fine-tuned iteratively for $40$ epochs.

\par The PyTorch framework was used for CNN training and feature extraction. The average of the scores from the multiple gallery images per test is used for the final identity verification. PyTorch Mobile\footnote{\url{https://pytorch.org/mobile/home/}} was utilized to deploy the models on-device. 

\section{Results}
Table~\ref{baseline} illustrates the performance of ResNet-50 fine-tuned on 200 subjects from VISOB datasets and tested on VISOB (same dataset) and UFPR-Periocular (cross dataset) under subject independent analysis. This model is used as a base (teacher) model for pruning (knowledge distillation) method.

\begin{table}[htb]
 \caption{Performance of ResNet-50 model fine-tuned on VISOB dataset and tested on VISOB~\cite{VISOB}, and UFPR-Periocular~\cite{zanlorensi2020ufprperiocular} datasets under subject independent analysis.This model is used as the base model of all the compression techniques.}
\label{baseline}

    \resizebox{\columnwidth}{!}{%
   \begin{tabular}{lllcccccc}
    \toprule
    \multicolumn{1}{c}{\multirow{2}[4]{*}{\textbf{Baseline CNN}}} &
    \multicolumn{1}{c}{\multirow{2}[4]{*}{\textbf{ Val.(\%)}}} & 
    \multicolumn{1}{c}{\multirow{2}[4]{*}{\textbf{Dataset}}} & 
    
     \multicolumn{1}{c}{\multirow{2}[4]{*}{\textbf{EER}}} & 
         \multicolumn{2}{c}{\textbf{GMR(\%) @ FMR}} & 
    \multicolumn{1}{c}{\multirow{2}[4]{*}{\textbf{AUC}}}\\
\cmidrule{5-6}       & & &    & \@\textbf{0.01} & \@\textbf{0.1} & &   \\
    \midrule
    
ResNet-50 & \quad 99.7\ & \quad VISOB\ &  8.22\ &	45.30	 \ & 54.74	 \ &	0.97	\ \\
& \quad & \quad UFPR\ &   6.84\ &	39.49	 \ &	58.73	 \ &	0.98	\ \\

    \bottomrule
    
    \end{tabular}}
\end{table}

The EERs of the model are $8.22\%$ and $6.84\%$ on VISOB and UFPR datasets (see Table~\ref{baseline}). The Genuine Match Rate (GMR) of the model is highest for VISOB at $45.30$ at False Match Rate (FMR) of $0.01$.  The Area Under Curve (AUC) of the model when evaluated on VISOB and UFPR dataset is almost equivalent. The equivalent performance on VISOB and UFPR-Periocular could be due to a low number of images per subject for UFPR-Periocular over VISOB.

\par Table~\ref{tab:prune_visob} shows the subject-independent verification performance analysis of pruning strategies evaluated using GMRs and EER on \textit{VISOB} dataset. The Layerwise Magnitude (LM) method obtained the lowest EER of $14.34\%$ at compression ratio (CR) of $8$ (which is $4\%$ increment over the baseline and $0.04$ decrease in AUC) and an average EER of $20.42\%$ over 6 compression ratios. The change in AUC of the pruned models over the AUC of ResNet-50 baseline is shown as well.

In comparison to LM, other strategies namely, Global Magnitude (GM), Global Gradient Magnitude (GGM), Layerwise Gradient Magnitude (LGM), and Random Pruning (RP) obtained $34.26\%$, $10.99\%$, $8.89\%$, and $6.53\%$ increase in EER, respectively. 

The standard deviation of intra-EERs across compression rates for each strategy is $23.0$\% for GM (being the highest), and for the rest falling in the range $[3.38-5.68]$. Surprisingly, the GM outperformed others at GMR scores, with a $14$-point increase @FPR=0.01 and $9.37$ @FPR=0.1 across all CRs. This increase in performance of GM-based pruning was not the case for UFPR, where it obtained an average EER of $86.58$\% with a gap of $63.24$\% to RP method on the UFPR dataset (shown in Table~\ref{tab:prune_upfr}). Overall, GM obtained the highest error rates even over RP and LM obtained the least error rate. Similarly, LM obtained the least error rates among others on \textit{UFPR} as well with an EER of $10.85\%$ at CR of $8$ which is $4\%$ increment over the baseline ($6.84\%$). The decrease in AUC over the baseline is $0.02$ for LW at CR of $8$.

Table~\ref{tab:kd_results} shows the subject-independent analysis of compact models (MobileNet-V2, MobileNet-V3, ResNet-8, ResNet-20, ShuffleNet-V2) obtained using knowledge distillation (KD) approach on VISOB and UFPR datasets. 
As can be seen, MobileNet-V2 outperformed other students (namely, ResNet-8, ResNet-20, MobileNet-V3) as it obtained $5.21\%$ EER, $49.09\%$ GMR@FMR=$0.01$\%, and $61.67\%$ GMR@FMR=$0.1$\% (as shown in Table~\ref{tab:kd_results}). The ResNet-20 had a slightly higher EER of $8.80\%$ over MobileNet-V2 and lower GMRs by a drop of $8.41$\% and $10.95$ at GMR@FMR=$0.01$ and $0.1$, respectively. However, MobileNet-V3 obtained the highest EER with an increase of $21.23$\% over the average of the other four students on both datasets. 
The EER of the MobileNet-V2 trained on VISOB without knowledge distillation is $8.0\%$.
The increment in AUC is $0.02$ over the baseline ResNet-50 model.
Similarly, for UPFR dataset, MobileNet-V2 outperformed other students (namely, ResNet-8, ResNet-20, MobileNet-V3) as it obtained $5.38\%$ EER, $50.19\%$ GMR@FMR=$0.01$\%, and $68.16\%$ GMR@FMR=$0.1$\%. The increment in AUC is $0.01$ over the baseline ResNet-50 model. 
MobileNet-V3 obtained the least performance on both datasets. The reason could be MobileNet-V3 dependence on the auto-search mechanism to find the best mobile architecture may not be effective for KD-based training. 

The second-lowest performance was obtained by ShuffleNet-V2 with an average EER of $23.0$\% and GMR@FPR=0.001 to $13.80$\%. ResNet-8 obtained better overall results over ShuffleNet-V2 while being $18.7$x smaller. The student MobileNet-V2 performed better than the teacher model after knowledge distillation by about $2.2$\% lower EER. This could be due to the sequential self-teaching of the student in better learning the inter-class similarity~\cite{Gou_2021}.  


\par Overall, \emph{experiments suggest the superiority of KD over other pruning methods in subject-independent evaluation}. This can be seen using Area Under the Curve (AUC) metric as the student networks obtained an average of $0.82$ AUC ($0.89$ for VISOB and UPFR) compared to $0.93$ obtained by the baseline model (ResNet-50). However, pruning methods obtained an average AUC of $0.72$ ($0.75$ and $0.69$ on VISOB and UFPR, individually).
Comparison of pruning methods across compression ratios suggests that the EER averaged across all CR points remained constant for the UFPR dataset i.e., $32.96$\%. The least performance was obtained at CR of 64 with an average EER value of $46.14$\% on VISOB and $43.54$\% on UFPR. At the same point, the average AUC on both datasets is $0.60$. However, \emph{the best performance was obtained for CR of 2} with an average EER of $21.35$\% and an average AUC of $0.87$.

\par To \emph{validate the latency of models} and to study the trade-off between performance and feature extraction speed, we measured the Extraction Time (ET) on real handheld devices as indicated in Table~\ref{tab:inf_time}. The aim of this experiment is not to compare the performance of different mobile devices but the real-time inferences of different compression techniques on different mobile platforms. Available Apple mobile devices were used as as they are the most adopted and popular. MobileNet-V3 was tuned to CPUs of mobile phones by the adaptation of Network Architecture Search (NAS) and complemented by the NetAdapt algorithm. This is evident as MobileNet-V3 being 1.17x faster than MobileNet-V2 across all five devices. The fastest ET was obtained by ResNet-8 with an average of $218$ms across all the devices, which makes it 6.3x faster and 341x smaller than ResNet-50 with an average ET of $1381$ ms. The average ET of $554$ms and $520.2$ ms were obtained for MobileNet-V2 and ResNet-20 models, respectively. We further calculated ET of the best-pruned model obtained using Layerwise Magnitude at CR of 2 for iPhone 6 and iPhone X. However, the ET for the pruned model averaged to be $1433$ ms (shown in Table~\ref{tab:inf_time}. Thus the pruned model ran 1.6x slower than the baseline model. 

\emph{One drawback of the unstructured pruning methods is that they result in having sparse weight matrices}, thus leading to inefficiency in speedup and compression on CPUs and GPUs, and also requiring having dedicated hardware~\cite{han2016eie}. Therefore, the pruned model did not obtain high efficiency during the run-time. \emph{The benefits of the pruning methods include reducing the total number of energy-intensive memory accesses and improving the inference time due to effectively higher memory bandwidth for fetching compressed model parameters. MobileNet-V2 and ResNet-20 student models offered the best trade-off between performance and speed when trained using KD with ResNet-50 as the teacher model}.

\begin{table}[htb]
 \caption{Subject-independent evaluation of the pruning methods at six \textbf{C}ompression \textbf{R}atios (CR) on VISOB Dataset~\cite{VISOB}. }
\label{my-label}
\resizebox{\columnwidth}{!}{%
   \begin{tabular}{llllccc}
    \toprule
    \multicolumn{1}{c}{\multirow{2}[4]{*}{\textbf{Pruning Method}}} & \multicolumn{1}{c}{\multirow{2}[4]{*}{\textbf{CR}}} &  \multicolumn{1}{c}{\multirow{2}[4]{*}{\textbf{EER}}} & 
    \multicolumn{2}{c}{\textbf{GMR(\%) @ FMR}} & 
    \multicolumn{1}{c}{\multirow{2}[4]{*}{\textbf{AUC}}}\\
\cmidrule{4-5}       &    &    & \@\textbf{0.01} & \@\textbf{0.1} & &   \\
    \midrule
   \qquad Global Magnitude (GM)& 
   \quad 2 & 22.36\ & 9.56\ & 13.28\ & 0.85 \(\downarrow \fpeval{0.97-0.85}\)\ \\
    &  \quad  4 & 29.72\ & 7.20\ & 22.27\ & 0.48 \(\downarrow \fpeval{0.97-0.48}\)\ \\
   
 \quad  & \quad 8 &  \ 53.69& 22.27\ & 22.27\ & 0.48 \(\downarrow \fpeval{0.97-0.48}\)\ \\
       & \quad 16 & 50.0\ & 2.50\ & 2.50\ & 0.50 \(\downarrow \fpeval{0.97-0.50}\)\ \\
       
  \qquad   & \quad 32  & 54.72\ & 15.09\ & 15.09\ & 0.44 \(\downarrow \fpeval{0.97-0.44}\)\ \\
       & \quad 64 & 100.0\ & 0.00\ & 0.00\ & 0.00 \(\downarrow \fpeval{0.97-0}\)\ \\

    \midrule
    
  \qquad Global Gradient Magnitude (GGM) & 
\quad 2 & 22.52\  & 10.98\ & 14.79\ & 0.85 \(\downarrow \fpeval{0.97-0.85}\)\ \\
    &  \quad  4 & 29.72\ & 7.27\ & 10.05\ & 0.78 \(\downarrow \fpeval{0.97-0.78}\)\ \\
   
 \qquad  & \quad 8 &  33.47\ & 6.18\ & 7.81\ & 0.74 \(\downarrow \fpeval{0.97-0.74}\)\ \\
       & \quad 16 & 28.81\ & 6.43\ & 9.86\ & 0.77 \(\downarrow \fpeval{0.97-0.77}\)\ \\
       
  \qquad   & \quad 32  & \ 29.91& 5.90\ & 8.24\ & 0.76 \(\downarrow \fpeval{0.97-0.76}\)\ \\
       & \quad 64 & 31.42\ & 5.70\ & 8.18\ & 0.74 \(\downarrow \fpeval{0.97-0.74}\)\ \\

    \midrule
       \qquad Layerwise Magnitude (LM) & 
   \quad 2 & 15.55\ & 16.88\ & 24.22\ & 0.92\(\downarrow \fpeval{0.97-0.92}\)\ \\
   \qquad  &  \quad  4 & 16.96\ & 15.07\ & 21.63\ & 0.91 \(\downarrow \fpeval{0.97-0.91}\)\ \\
   
 \qquad  & \quad 8 &  14.34\ & 19.51\ & 27.45\ & 0.93 \(\downarrow \fpeval{0.97-0.93}\)\ \\
       & \quad 16 & 23.54\ & 9.29\ & 12.80\ & 0.84 \(\downarrow \fpeval{0.97-0.84}\)\ \\
       
  \qquad   & \quad 32  & 22.43\ & 11.15\ & 15.30\ & 0.85 \(\downarrow \fpeval{0.97-0.85}\)\ \\
       & \quad 64 & 29.71\ & 6.02\ & 8.02\ & 0.76 \(\downarrow \fpeval{0.97-0.76}\)\ \\

    \midrule
       \qquad Layerwise Gradient Magnitude (LGM) & 
   \quad 2 & 26.01\ & 7.56\ & 10.86\ & 0.83 \(\downarrow \fpeval{0.97-0.83}\)\ \\
    &  \quad  4 & 26.12\ & 8.24\ & 11.02\ & 0.82 \(\downarrow \fpeval{0.97-0.82}\)\ \\
   
 \qquad  & \quad 8 &  \ 43.00& 1.42\ & 1.42\ & 0.59 \(\downarrow \fpeval{0.97-0.59}\)\ \\
       & \quad 16 & 31.29\ & 4.36\ & 4.94\ &0.75 \(\downarrow \fpeval{0.97-0.75}\)\ \\
       
  \qquad   & \quad 32  & 30.40\ & 7.16\ & 9.12\ & 0.77 \(\downarrow \fpeval{0.97-0.77}\)\ \\
       & \quad 64 & 31.85\ & 4.39\ & 4.83\ & 0.73 \(\downarrow \fpeval{0.97-0.73}\)\ \\
    \midrule
        \qquad Random Pruning (RP) & 
   \quad 2 & 22.40\ & 9.97\ & 13.68\ & 0.86 \(\downarrow \fpeval{0.97-0.86}\)\ \\
   &  \quad  4 & 26.08\ & 7.41\ & 10.00\ & 0.82 \(\downarrow \fpeval{0.97-0.82}\)\ \\
   
 \qquad  & \quad 8 &  23.38\ & 10.79\ & 14.63\ & 0.84 \(\downarrow \fpeval{0.97-0.84}\)\ \\
       & \quad 16 & 25.74\ & 6.75\ & 10.76\ & 0.82 \(\downarrow \fpeval{0.97-0.82}\)\ \\
      
  \qquad  & \quad 32  & 29.10\ & 5.52\ & 7.62\ & 0.77 \(\downarrow \fpeval{0.97-0.77}\)\ \\
       & \quad 64 & 35.00\ & 2.81\ & 4.20\ & 0.70 \(\downarrow \fpeval{0.97-0.70}\)\ \\

    \bottomrule
    \end{tabular}}
    \label{tab:prune_visob}
\end{table}

\begin{table}[htb]
 \caption{Subject-independent evaluation of the pruning methods at six \textbf{C}ompression \textbf{R}atios (CR) on UFPR Dataset~\cite{zanlorensi2020ufprperiocular}.}
\label{my-label}
\resizebox{\columnwidth}{!}{%
   \begin{tabular}{llllccc}
    \toprule
    \multicolumn{1}{c}{\multirow{2}[4]{*}{\textbf{Pruning Method}}} & \multicolumn{1}{c}{\multirow{2}[4]{*}{\textbf{CR}}} &  \multicolumn{1}{c}{\multirow{2}[4]{*}{\textbf{EER}}} & 
    \multicolumn{2}{c}{\textbf{GMR(\%) @ FMR}} & 
    \multicolumn{1}{c}{\multirow{2}[4]{*}{\textbf{AUC}}}\\
\cmidrule{4-5}       &    &    & \@\textbf{0.01} & \@\textbf{0.1} & &   \\
    \midrule
  \qquad Global Magnitude (GM) & 
   \quad 2 & 19.46\ & 9.54\ & 19.19\ & 0.88 \(\downarrow \fpeval{0.98-0.88}\)\ \\
    &  \quad  4 & 100.0\ & 0.00\ & 0.00\ & 0.00 \(\downarrow \fpeval{0.98-0.0}\)\ \\
   
 \quad  & \quad 8 & 100.0\ & 0.00\ & 0.00\ & 0.00 \(\downarrow \fpeval{0.98-0.0}\)\ \\
       & \quad 16 & 100.0\ & 0.00\ & 0.00\ & 0.00 \(\downarrow \fpeval{0.98-0.0}\)\ \\
       
  \qquad   & \quad 32 & 100.0\ & 0.00\ & 0.00\ & 0.00 \(\downarrow \fpeval{0.98-0.0}\)\ \\
       & \quad 64 & 100.0\ & 0.00\ & 0.00\ & 0.00 \(\downarrow \fpeval{0.98-0.0}\)\ \\

    \midrule
    
   \qquad Global Gradient Magnitude (GGM)& 
\quad 2 & 21.41\  & 2.85\ & 7.46\ & 0.86 \(\downarrow \fpeval{0.98-0.86}\)\ \\
    &  \quad  4 & 24.87\ & 3.06\ & 6.94\ & 0.83 \(\downarrow \fpeval{0.98-0.83}\)\ \\
   
 \qquad  & \quad 8 &  29.79\ & 3.15\ & 8.76\ & 0.78 \(\downarrow \fpeval{0.98-0.78}\)\ \\
       & \quad 16 & 26.50\ & 5.25\ & 12.50\ & 0.82 \(\downarrow \fpeval{0.98-0.82}\)\ \\
       
  \qquad   & \quad 32  & 28.23\ & 6.50\ & 13.83\ & 0.79 \(\downarrow \fpeval{0.98-0.79}\)\ \\
       & \quad 64 & 29.10\ & 4.68\ & 10.32\ & 0.78 \(\downarrow \fpeval{0.98-0.78}\)\ \\

    \midrule
       \qquad Layerwise Magnitude (LM) & 
   \quad 2 & 14.10\ & 13.11\ & 28.06\ & 0.93 \(\downarrow \fpeval{0.98-0.93}\)\ \\
   \qquad  &  \quad  4 & 12.34\ & 19.93\ & 36.69\ & 0.95 \(\downarrow \fpeval{0.98-0.95}\)\ \\
   
 \qquad  & \quad 8 &  10.86\ & 23.37\ & 38.64\ & 0.96 \(\downarrow \fpeval{0.98-0.96}\)\ \\
       & \quad 16 & 17.68\ & 15.34\ & 27.03\ & 0.91 \(\downarrow \fpeval{0.98-0.91}\)\ \\
       
  \qquad   & \quad 32  & 18.49\ & 14.92\ & 27.84\ & 0.89 \(\downarrow \fpeval{0.98-0.89}\)\ \\
       & \quad 64 & 30.48\ & 8.13\ & 15.19\ & 0.75 \(\downarrow \fpeval{0.98-0.75}\)\ \\

    \midrule
       \qquad Layerwise Gradient Magnitude (LGM) & 
   \quad 2 & 27.84\ & 4.32\ & 10.30\ & 0.80 \(\downarrow \fpeval{0.98-0.80}\)\ \\
    &  \quad  4 & 26.41\ & 2.39\ & 5.49\ & 0.81 \(\downarrow \fpeval{0.98-0.81}\)\ \\
   
 \qquad  & \quad 8 &  40.16\ & 0.730\ & 2.01\ & 0.63 \(\downarrow \fpeval{0.98-0.63}\)\ \\
       & \quad 16 & 28.25\ & 4.56\ & 10.51\ &0.79 \(\downarrow \fpeval{0.98-0.79}\)\ \\
       
  \qquad   & \quad 32  & 24.81\ & 7.14\ & 14.92\ & 0.82 \(\downarrow \fpeval{0.98-0.82}\)\ \\
       & \quad 64 & 26.81\ & 4.98\ & 10.80\ & 0.80 \(\downarrow \fpeval{0.98-0.80}\)\ \\
    \midrule
      \qquad Random Pruning (RP) & 
   \quad 2 & 21.61\ & 0.951\ & 7.41\ & 0.86 \(\downarrow \fpeval{0.98-0.86}\)\ \\
   &  \quad  4 & 21.39\ & 7.18\ & 16.62\ & 0.86 \(\downarrow \fpeval{0.98-0.86}\)\ \\
   
 \qquad  & \quad 8 &  19.70\ & 10.10\ & 24.11\ & 0.88 \(\downarrow \fpeval{0.98-0.88}\)\ \\
       & \quad 16 & 20.92\ & 11.00\ & 20.92\ & 0.87 \(\downarrow \fpeval{0.98-0.87}\)\ \\
       
  \qquad   & \quad 32  & 25.19\ & 8.37\ & 14.99\ & 0.83 \(\downarrow \fpeval{0.98-0.83}\)\ \\
       & \quad 64 & 31.32\ & 4.79\ & 9.30\ & 0.74 \(\downarrow \fpeval{0.98-0.74}\)\ \\

    \bottomrule
    \end{tabular}}
    \label{tab:prune_upfr}
\end{table}

\begin{table}[htb]
 \caption{Subject independent analysis of the compact models generated using knowledge distillation method with ResNet-50 as the teacher model on VISOB~\cite{VISOB} and UFPR~\cite{zanlorensi2020ufprperiocular} datasets.}
\label{my-label}

    \resizebox{\columnwidth}{!}{%
   \begin{tabular}{lllcccccc}
    \toprule
    \multicolumn{1}{c}{\multirow{2}[4]{*}{\textbf{Dataset}}} &
    \multicolumn{1}{c}{\multirow{2}[4]{*}{\textbf{Student}}} & 
    \multicolumn{1}{c}{\multirow{2}[4]{*}{\textbf{ Val.}}} & 
     \multicolumn{1}{c}{\multirow{2}[4]{*}{\textbf{EER}}} & 
         \multicolumn{2}{c}{\textbf{GMR(\%) \@ FMR}} & 
    \multicolumn{1}{c}{\multirow{2}[4]{*}{\textbf{AUC}}}\\
\cmidrule{5-6}       & & &    & \@\textbf{0.01} & \@\textbf{0.1} & &   \\
    \midrule
    
VISOB & \quad ResNet-8& \quad 59\ &  12.83   \ &    34.30    \ &42.48    \ &    0.94 \(\downarrow \fpeval{0.97-0.94}\)\ \\
& \quad ResNet-20& \quad 84\ &       8.80\ &    40.64    \ &    50.72    \ &    0.97 \(\downarrow \fpeval{0.97-0.97}\)\ \\

& \quad MobileNet-V2& \quad 99\ &  5.21  \ &    49.09    \ &    61.67    \ &    0.99 \(\downarrow \fpeval{0.99-0.97}\)\ \\

& \quad MobileNet-V3& \quad 99.5\ & 34.80\ &    4.47     \ &    6.75     \ &    0.71 \(\downarrow \fpeval{0.97-0.71}\)\ \\

& \quad ShuffleNetV2-50 & \quad 98\ &  23.41     \ &    16.19    \ &    21.36    \ &    0.84 \(\downarrow \fpeval{0.97-0.84}\)\ \\

\midrule
UFPR & \quad ResNet-8& \quad 59\ & 15.83     \ &    22.53    \ &    36.61    \ &    0.92 \(\downarrow \fpeval{0.98-0.92}\)\ \\

& \quad ResNet-20& \quad 84\ & 10.98    \ & 34.78    \ &    50.38    \ &    0.95 \(\downarrow \fpeval{0.98-0.95}\)\ \\

& \quad MobileNet-V2& \quad 99 \ & 5.38  \ &    50.19    \ &    68.16    \ &    0.99 \(\downarrow \fpeval{0.99-0.98}\)\ \\

& \quad MobileNet-V3& \quad 99.5\ & 33.97    \ &    2.73     \ &    6.35     \ &    0.72 \(\downarrow \fpeval{0.98-0.72}\)\ \\

& \quad ShuffleNetV2-50 & \quad 98\ & 22.52  \ &    11.38    \ &    22.00    \ &    0.85 \(\downarrow \fpeval{0.98-0.85}\)\ \\

    \bottomrule
    
    \end{tabular}}
    \label{tab:kd_results}
\end{table}

\begin{table}[htb]
 \caption{\textbf{Inference time}: Feature \textbf{E}xtraction \textbf{T}ime (ET) of all the compact models in milliseconds (ms).}
\label{my-label}
\resizebox{\columnwidth}{!}{%
   \begin{tabular}{lllcc}
    \toprule
      \multicolumn{1}{c}{{\textbf{CNN}}} &  \multicolumn{1}{c}{\textbf{Size (MB)}} 
      &  \multicolumn{1}{c}{\textbf{\# Params (m)}}
      & \multicolumn{1}{c}{\textbf{Device}}
      & \multicolumn{1}{c}{\textbf{ET (ms)}}
      \\

    \midrule
    \qquad ResNet-50& \quad 98.8 & \quad 25.6 & iPhone 6 & 2386\  \\
                         & & &iPhone X & 941\  \\
                         & & &iPhone XR & 957\  \\
                         & & &iPad Air 2 & 1676\  \\
                         & & &iPad 7th Gen.& 945\  \\

     \midrule
    \qquad ResNet-20& \quad 1.3  & \quad 2.7& iPhone 6 & 1261\  \\
                         & & &iPhone X & 278\  \\
                         & & &iPhone XR & 185\  \\
                         & & &iPad Air 2 & 585\  \\
                         & & &iPad 7th Gen.& 292\  \\
      \midrule
    \qquad ResNet-8& \quad 0.49 & \quad 0.075 & iPhone 6 & 533\  \\
                         & & &iPhone X & 120\  \\
                         & & &iPhone XR & 75\  \\
                         & & &iPad Air 2 & 237\  \\
                         & & &iPad 7th Gen.& 124\  \\
                         
          \midrule
    \qquad MobileNet-V2& \quad 12 & \quad 3.5& iPhone 6 & 1022\  \\
                         & & &iPhone X & 357\  \\
                         & & &iPhone XR & 243\  \\
                         & & &iPad Air 2 & 774\  \\
                         & & &iPad 7th Gen.& 374\  \\
       \midrule   
   \qquad MobileNet-V3& \quad 8.6 & \quad  2.5& iPhone 6 & 797\  \\
                 & & &iPhone X & 314\  \\
                 & & &iPhone XR & 205\  \\
                 & & &iPad Air 2 & 700\  \\
                 & & &iPad 7th Gen.& 352\  \\
\midrule                  
    \qquad ShuffleNetV2-50& \quad 3.9  & \quad  1.4 & iPhone 6 & 858\  \\
                     & & &iPhone X & 299\  \\
                     & & &iPhone XR & 205\  \\
                     & & &iPad Air 2 & 712\  \\
                     & & &iPad 7th Gen.& 358\  \\
                     
 \midrule                  
    \qquad Pruned & \quad 93.70   & \quad  24.6  & iPhone 6 & 3908\  \\
    \qquad                 & & &iPhone X & 1431\  \\

    \bottomrule
    \end{tabular}}
    \label{tab:inf_time}
\end{table}

\section{Conclusion}
This study evaluates five neural network pruning algorithms for CNN-based on-device smartphone user authentication based on ocular images for the first time. Comparative analysis is performed with knowledge distillation methods on publicly available VISOB and UFPR periocular datasets.  Experimental results suggest the efficacy of knowledge distillation-based methods over network pruning on an average. Specifically, MobileNet-V2 and ResNet-20 based student models trained using knowledge distillation with ResNet-50 as the teacher obtained the best tradeoff between accuracy and speed when evaluated on five mobile devices. Thus, demonstrating the efficacy of knowledge distillation over other pruning techniques for real-time smart-phone user authentication on mobile devices using ocular images. 
As a part of future work, patch-based CNN and those based on structured pruning techniques such as layer removal will be evaluated and compared on the same test bed.\\




{\small
\bibliographystyle{ieee}
\bibliography{sample, sample1,mybibliography}
}



\end{document}